# Polycraft World AI Lab (PAL) : An Extensible Platform for Evaluating Artificial Intelligence Agents

Stephen A. Goss[1,*], Robert J. Steininger[1,*], Dhruv Narayanan[1], Daniel V. Olivença[2], Yutong Sun[2], Peng Qiu[2], Jim Amato[1], Eberhard O. Voit[2], Walter E. Voit[1,3], Eric J. Kildebeck[1,**]

[1] Center for Engineering Innovation, The University of Texas at Dallas, 800 W. Campbell Road, Richardson, Texas 75080, USA
[2] Department of Biomedical Engineering, Georgia Institute of Technology and Emory University, Atlanta, GA, United States
[3] Department of Materials Science and Engineering, The University of Texas at Dallas, 800 W. Campbell Road, Richardson, Texas 75080, USA

Emails:

stephen@polycraftworld.com

robert.steininger@utdallas.edu

dhruv@polycraftworld.com

dolivenca3@gatech.edu

sunyutong@gatech.edu

peng.qiu@bme.gatech.edu

Jim@polycraftworld.com

eberhard.voit@bme.gatech.edu

walter.voit@utdallas.edu

eric.kildebeck@utdallas.edu

*Co-first authors

**Corresponding author: University of Texas at Dallas, 800 W. Campbell Road, Richardson, Texas 75080, USA

# Abstract

As artificial intelligence research advances, the platforms used to evaluate AI agents need to adapt and grow to continue to challenge them. We present the Polycraft World AI Lab (PAL), a task simulator with an API based on the Minecraft mod Polycraft World. Our platform is built to allow AI agents with different architectures to easily interact with the Minecraft world, train and be evaluated in multiple tasks. PAL enables the creation of tasks in a flexible manner as well as having the capability to manipulate any aspect of the task during an evaluation. All actions taken by AI agents and external actors (non-player-characters, NPCs) in the open-world environment are logged to streamline evaluation. Here we present two custom tasks on the PAL platform, one focused on multi-step planning and one focused on navigation, and evaluations of agents solving them. In summary, we report a versatile and extensible AI evaluation platform with a low barrier to entry for AI researchers to utilize.

# Introduction

Advances in artificial intelligence algorithms, in particular advances in open world learning and lifelong learning agents, require scalable open-world testbeds for development and evaluation (Doctor et al., 2022). Here, we present a Minecraft-based platform for simulating a wide variety of complex tasks to evaluate the next generation of artificial intelligence agents.

Minecraft is a video game that simulates a rich 3D environment where players collect resources and blocks to build anything from rich urban environments to virtual computing systems. Further, Minecraft has an extensive modding community where the bounds of the base game can be extended. We created a comprehensive, fully customizable mod to Minecraft, Polycraft World (Smaldone et al., 2017) initially to teach materials science and polymer chemistry. The Polycraft World mod introduces over 5,000 new materials, items, and machinery on top of base Minecraft. In addition to increased complexity of items, this mod introduces hierarchal player permissions, player/agent tracking and logging. Through this continued development, Polycraft World has grown to a multipurpose platform with many applications from education to social science to artificial intelligence.

To evaluate open-world learning, transfer learning, and lifelong learning, an ideal virtual platform for evaluating AI agents should support a variety of tasks while still being broadly usable for different agent types. Many platforms are limited to a task or tasks within a discrete, closed environment (e.g., CartPole (Sutton & Barto, 2018), Arcade Learning Environment (Bellemare et al., 2013)). Virtual 3D worlds provide added value in that they can better represent real-world relevant tasks and complex scenarios. The Unity game engine can be a powerful platform for AI evaluation (Juliani et al., 2018); however, there is a high development cost to create custom Unity based tasks. Additionally, many first-person shooter games have been adapted into AI research platforms such as VizDoom (Doom) (Kempka et al., 2016) and DeepMind Lab (Beattie et al., 2016) using the Quake III Arena game engine. ViZDoom and DeepMind Lab are excellent platforms to train AI agents in a 3D visual environment, but they are essentially first-person shooters with limited interactions with the environment often focused on movement with a single or small number of interaction/fire commands. While more interactions can be added to these platforms, Minecraft presents a richer out-of-the-box experience. Minecraft has been used in the past in this capacity as well (Aluru et al., 2015; Johnson et al., 2016). Previous Minecraft platforms have been invaluable for studying and challenging AI agents. However, they are limited by the bounds of Minecraft itself. Tasks have limited flexibility, and agents must be adapted to function within their system.

In this paper, we present the Polycraft World AI Lab (PAL), which retains the advantages of Minecraft-based platforms but has the added advantage of including an API well-suited for planning agents that creates a complex action space without requiring extensive overhead training for movement or reinforcement learning. With this API, PAL enables more flexible tasks that are straightforward to customize and are more approachable for various AI agents. We first describe the platform architecture. We then explain how one would design and implement a task on PAL. We go on to present the implementation of two custom tasks. Finally, we discuss the PAL outputs for evaluation and platform performance metrics.

## PAL Architecture

The Polycraft World AI Lab (PAL) platform (https://github.com/PolycraftWorld/PAL) was built to test AI agents on top of Polycraft World (Figure 1) using the popular modding API, Minecraft Forge (Forge volunteer team, 2011). The PAL platform consists of the Minecraft Client and Polycraft World mod connected via the Minecraft Forge API, a Web Socket API, and a Tournament/Game manager for handling experiments with different complex tasks to test AI agents. Step by step instructions for PAL installation can be found in the GitHub page for Linux and Windows but any system that can use Java will be able to run PAL. Our agent API allows interfacing with any external AI agent without specific language requirements. All that is needed to integrate a new agent is a simple socket client script, which can be made in any language that supports web sockets such as Java, Python, and Julia. This significantly lowers the barrier for researchers to utilize the PAL platform as they will not need to develop or integrate with an existing wrapper. The Tournament/Game manager is our system for facilitating evaluation of AI agents in custom PAL task scenarios. A more detailed explanation of the PAL architecture can be found in the Supplemental Materials, in the "1. Process Communication Standards" section.

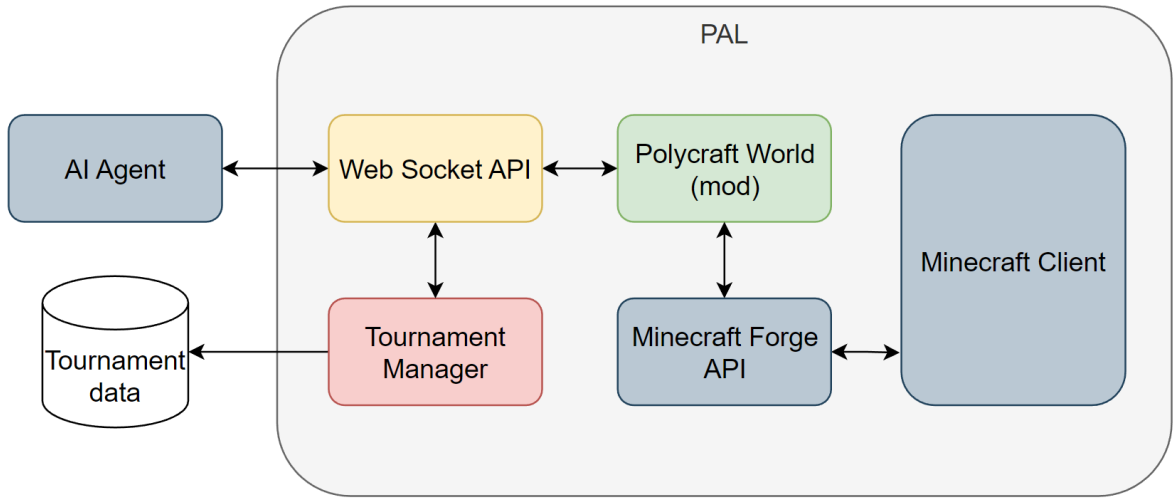

*Figure 1 : PAL Architecture Flow Chart*

**Interface - Actions**

Agents interact with the world by sending commands to PAL one command per game tick. The available commands (Table 1 and Supplemental Materials) are universal and do not change between tasks. There are different categories of commands such as game commands (START, RESET, GIVEUP), movement (MOVE, TP_TO), interactions (CRAFT, BREAK_BLOCK, USE_HAND, SELECT_ITEM), and sensing (SENSE_ALL, SENSE_SCREEN).  Each command has set preconditions that must be satisfied to execute. For example, an agent cannot move forward if there is a block in the way. Each command has an associated cost that can be used as a metric, for example, in reinforcement learning. The cost per action is based on an estimate of the complexity of the action. For example, Teleporting (TP_TO) 10 blocks costs 10X the cost of moving (MOVE) 1 block and the cost of crafting increases based on the number of items used in a recipe. The reasonable estimation of these costs allows for evaluation of agent efficiency and realism of actions in large open-world tasks.

Table 1: Action commands available in PAL.

| Interactions | Sensing | Movement | Game commands |
|---|---|---|---|
| BREAK_BLOCK | SENSE_SCREEN | MOVE | START |
| PLACE | SENSE_ALL | TURN | RESET |
| COLLECT | SENSE_ALL_NONAV | TILT | GIVE_UP |
| CRAFT | SENSE_INVENTORY | TP_TO (teleport to) | |
| SELECT_ITEM | SENSE_LOCATIONS | | |
| USE_HAND | SENSE_ACTOR_ACTIONS | | |

| | |
|---|---|
| USE | SENSE_RECIPES |
| DELETE | SENSE_ENTITIES |
| INTERACT | |
| TRADE | |
| NOP (no operation) | |

For navigation, we support discrete steps by moving one block at a time or by teleporting to block positions for agents that don't have the capability to navigate on their own. By using the MOVE command, an agent can move to any open adjacent square.

**Interface – Observations**

Agents can sense multiple observations about the world including symbolic observations (recipes, locations, inventory, map, actor actions, and entities) and visual observation of the screen. There are three higher level observation commands that return specific sets of these observations based on if the agent is using vision, is teleporting or navigating, etc. These commands are:

- SENSE_SCREEN that returns a .PNG image file with the output of the visual screen.
- SENSE_ALL will return a JSON string with all the information about the current state of the environment, which includes a symbolic map of the task's terrain with block and NPC information, inventory, NPC locations and actions, but will not include the screen information.
- SENSE_ALL NONAV is similar to the previous command, but the symbolic map also includes all attributes of each block in the world.

The following commands allow the agent to access specific parts of SENSE_ALL.

- SENSE_INVENTORY will return a JSON string with the agent's inventory contents.
- SENSE_LOCATIONS will return a JSON string with the agent's position.
- SENSE_ACTOR_ACTIONS will return a JSON string with all the actions NPCs have performed.
- SENSE_RECIPES will return a JSON string with all the available recipes in the CRAFT command.
- SENSE_ENTITIES will return a JSON string with the map coordinates of all NPCs.

**Interface – Example of Command and Response**

An agent can utilize the API by connecting to the default port 9000 and sending a command in plain text. Then an agent could issue a command by sending "SELECT_ITEM

minecraft:iron_pickaxe" to that port followed by a new line character. After the command is processed, a text JSON response will be sent back which includes the result of the command and task goal information:

```
{
  "goal": {
    "goalType": "ITEM",
    "goalAchieved": false,
    "Distribution": "Uninformed"
  },
  "command_result": {
    "command": "select_item",
    "argument": "minecraft:iron_pickaxe",
    "result": "SUCCESS",
    "message": "selected item",
    "stepCost": 120
  },
  "step": 0,
  "gameOver": false
}
```

An example Python script can be found in the Supplemental Materials in the "2 .Example of Command and Response" section.

**Logging**

The PAL platform records logs in three different tracks which helps facilitate data analysis and the debugging process. These logs are separated by the tournament manager, Minecraft client, and lastly the agent. Both the tournament manager and Minecraft logs will be the same for each evaluation; however, the agent log always records directly from the AI agent standard output, which can be very different for each agent. These log files, produced by PAL, provide an out of the box option for player/agent tracking and analytics.

**Task Scenario**

One strength of the PAL platform is its ability to support a variety of task scenarios which can be researcher-defined and have broad customizability. To define a PAL task scenario, one must set a goal, define an arena and a set of available actions and interactions. A goal is what the agent must accomplish to succeed in the task. This can be anything that a player could do in Polycraft, whether it be obtaining/crafting an item, placing a block in a specified location, or building a complex system/structure in the arena. The arena is where the agent will be acting to achieve the goal. We can customize the arena by making it larger/smaller, filling it with objects to help or hinder the agent and adding adversarial or allied external agents. The action space of a given task will be dependent on all the objects in the arena as well as items in the player's inventory. Task components are defined using the open standard file format, JSON (JavaScript Object Notation), to allow for effortless and scalable customization of training environments in a familiar programing format. For new tasks, Java scripts are used to generate new task definitions, which allows quick generation of multiple seeded variations of the scenario. For tasks that resemble existing tasks, this can be done by manually editing an existing task definition file. Below, we present two custom task scenarios used in the DARPA SAIL-ON program (Senator, 2019) to illustrate the types of tasks that can be created in PAL.

## Example Task Scenarios

A wide variety of tasks for AI agent evaluation can be developed utilizing the PAL platform. Here, we present two sample task scenarios created within our PAL platform and demonstrate evaluation of AI solutions for these tasks. The first task, called POGO, is a planning task where agents must obtain resources and ultimately craft a pogo stick. The second task is a navigation task named HUGA (for hunter-gatherer) where agents must find an object, pick it up, and bring it to a target. These tasks highlight strengths in our Minecraft-based platform. In Polycraft World rich virtual worlds can be created that, like the real world, can be challenging to navigate, are only partially observable, and in which the optimal course of action may be unclear.

PAL Task Scenario 1: POGO Task

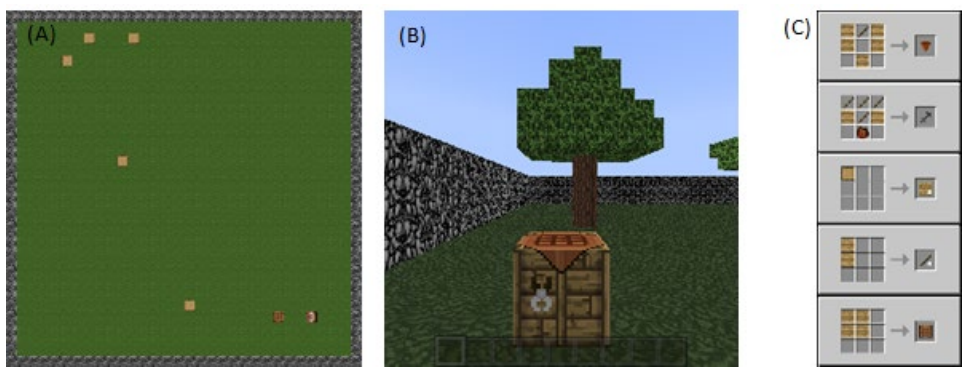

*Figure 2 : POGO Task. (A) Schematic of the POGO arena. (B) First person view of the agent within the POGO arena. (C) Recipes available to the agent.*

The goal of this task is to craft a pogo stick. The agent starts in a 30 x 30 block grass field surrounded by unbreakable bedrock walls (Figure 2A,B). The field always has 5 trees and 1 crafting table placed in the environment. Trees can be "chopped down" using the BREAK_BLOCK command, new items can be crafted from the available recipes (Figure 2C) utilizing CRAFT commands and the crafting table. Once the agent successfully obtains a pogo stick, the goal is achieved, and the task ends. The task can also end if the agent gives up or if the time limit for the task is reached.

To achieve the goal the agent must follow a series of steps including 1) chopping down trees to collect wood, 2) crafting intermediate items including wood planks, sticks, and a tree tap, 3) placing a tree tap on a tree, 4) collecting rubber from the tree tap, and 5) using intermediate items and rubber to craft a pogo stick.

We developed a planner agent for this task using Planning Domain Definition Language (PDDL) (Barrett et al., 1998; Fox & Long, 2003) and the fast forward algorithm (Hoffmann & Nebel, 2011). The planner uses the PDDL description of the task and fast forward to get a solution which will be used to create a sequence of actions that are necessary to achieve the task goal. It requires knowledge of the action space and the items available in the map.  The task components are translated into to PDDL 2.1 in the form of two files: the "domain.pddl" file that sets the world rules and the "problem.pddl" file that describes the task. The PDDL files construction is not automated but the agent is able to adapt to differences in the map composition. Running these files through the fast forward algorithm outputs a sequence of high-level actions that are interpreted by the Python code. High level actions aggregate low-level actions that Minecraft can process. As an example, a high-level action is 'get wood block' and the low-level actions corresponding to it are: 'move three steps forward', 'turn left', 'break wood block'. For this planner agent we use high-level actions to reduce the action space and to allow for adaptability to changes in the task (ex: crafting table and trees changing positions in the map). This agent does not require training.

PAL Task Scenario 2: Hunter-Gatherer Game (HUGA) – 3D space navigation task

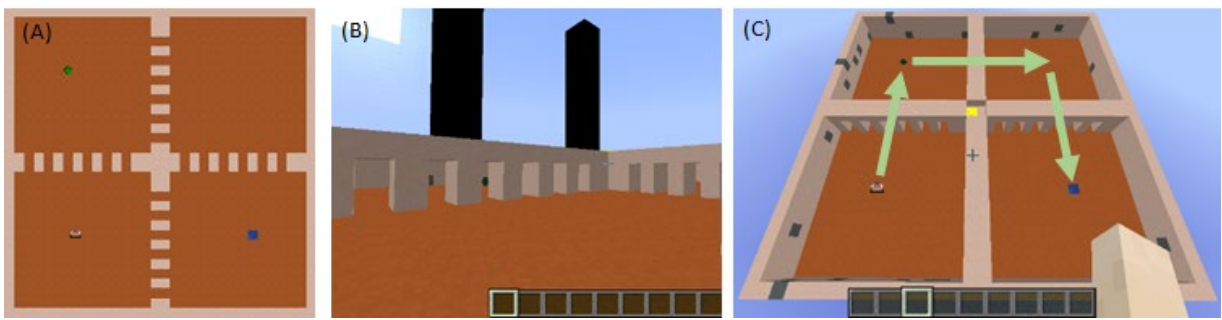

*Figure 3: Hunter Gatherer Task (HUGA) (A) Schematic view of the HUGA arena. (B) First person view of the agent within the HUGA arena. (C) Aerial view of the HUGA arena. Arrows indicate a path agent can take to complete the task*

The goal of this task is to pick up the green cross object (the MacGuffin) and place it on a Target (the blue block). The agent starts in a four-room arena with a total size of 32 x 32 blocks (Figure 3). Each room is 14 x 14 blocks large, with inner walls and multiple walkways separating the rooms. The agent always spawns in the bottom-left room (Room 1). The MacGuffin spawns in the top-left room (Room 2). The top-right room is empty (Room 3). The Target spawns in the bottom-right room (Room 4). The agent can pick up the MacGuffin by walking into it with MOVE commands and can place the MacGuffin at the Target with the PLACE command. Once the agent successfully places the MacGuffin at the Target, the goal is achieved and the task ends. The task can also end if the agent gives up or if the time limit is reached. This task is designed to be similar to other state-of-the-art 3D maze-runner tasks (e.g. VizDoom, DMLab NavMaze, Unity Obstacle Tower Challenge).

To achieve the goal the agent must follow a series of steps (Figure 3C) including 1) walking from Room 1 to Room 2, 2) picking up the MacGuffin, 3) walking from Room 2 to Room 4 and 4) placing the MacGuffin at the Target.

Like the POGO task, an agent was created to solve the HUGA task. Our HUGA agent uses a Deep Q-Network (DQN) (Mnih et al., 2013; Roderick et al., 2017) which combines deep neural networks and Q-learning. We chose to use this architecture because the deep neural network can take visual information as input, and the Q-learning aspect can handle the sparse reward structure in the HUGA task.

The structure of the DQN is determined by the sizes of the state space and the action space. A state for the visual-information-based DQN is the screenshot image of the game window at a specific time point. The dimension of the input is (3, 84, 84), where 3 is the number of color channels and (84, 84) is the resolution of the screenshot. The action space utilized by this agent includes MOVE, TURN, and PLACE for the MacGuffin (green block in Figure 3B,C). The DQN is trained to take the current state as input, and computes the expected rewards in the action space. After the agent executes the action with the highest expected reward, the new state is fed into the DQN to generate the next action. The process iterates until the agent completes the game.

To complete the HUGA game, the agent needs to find the MacGuffin and bring it to the destination (blue square in Figure 3A,C). This task can be separated into two sub-tasks. The first sub-task is to find the MacGuffin and the second sub-task is to find the destination. We trained two DQNs with identical architecture to complete the two sub-tasks.

More specifically, we generated 1000 instances of the HUGA task with different layouts of the passages among rooms and various color patterns on the walls and the ground. We randomly selected 618 instances for training, and used the remaining 382 for testing. For each training instance of the HUGA task, we randomly initialized the location and the direction the agent is facing (Figure 3B). We set the maximum number of actions to be 450. When the agent completed the sub-task within 450 actions or reaches 450 actions before completing the sub-task, the game was reset using the next training instance of the HUGA game, and training process continued.

When implementing the agent, we check the status after each action. Before finding the MacGuffin, the agent calls the first DQN model to decide its actions. After the MacGuffin is found, the agent turns to the second DQN model to decide its actions.

When the agent is tested in one instance of the HUGA task, we record a success if each sub-task is completed within 450 steps. We use the success rate as the evaluation metric for the trained agent. We randomly selected HUGA task instances from the 382 for testing, randomly set the location and direction the agent is facing, and tested the trained agent. We observed a success rate of ~94% with this agent.

## PAL Extensibility

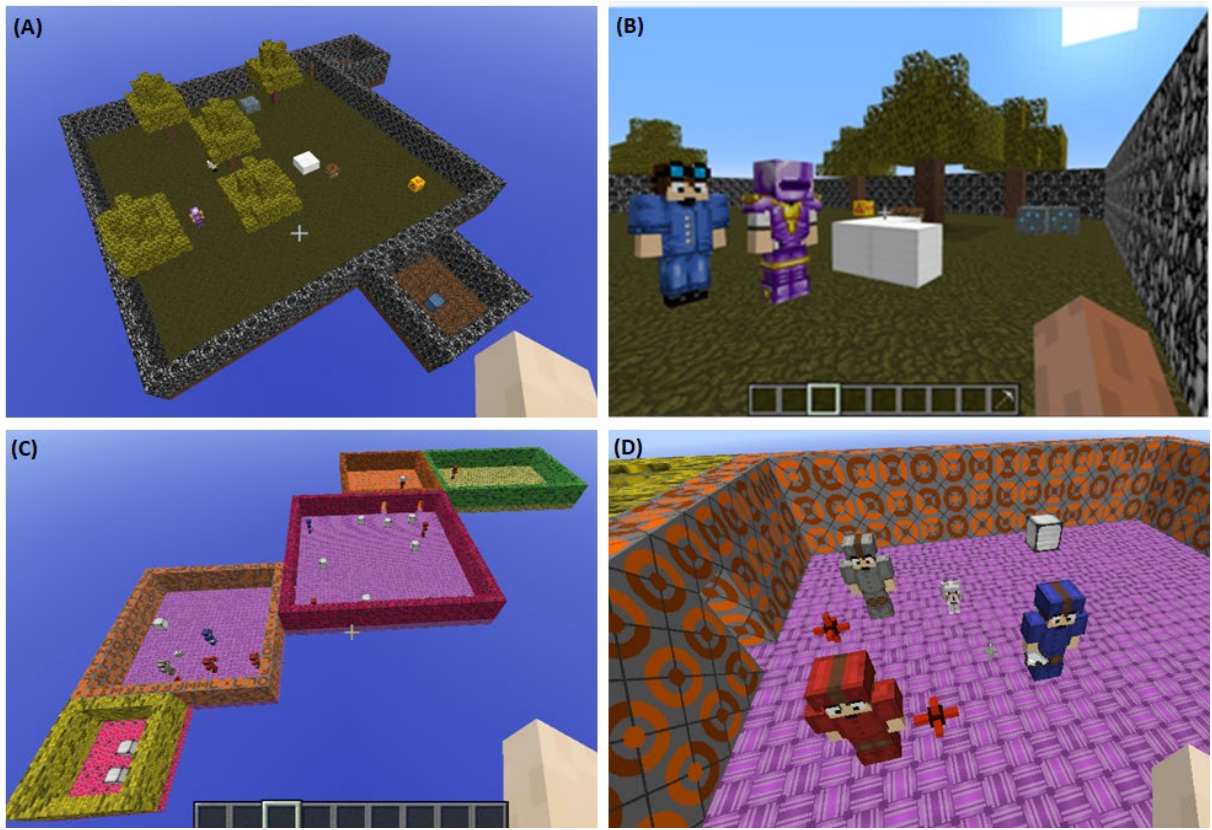

*Figure 4 – POGO and HUGA modifications (A) Bird's eye view of updated POGO arena (B) First person view of updated POGO task scenario (C) Bird's eye view of updated HUGA arena (D) First person view of updated HUGA task scenario.*

In Figure 4 we show modifications we've made to both the POGO and HUGA tasks that demonstrate the extensibility of our platform. Scale and complexity of tasks can be easily increased and decreased using our flexible task creation system based on the needs of AI researchers and the capability of their agents. We've increased the complexity of the POGO and HUGA tasks by adding more varied layouts, more complex problems, and external actors (NPCs), which can be allied or adversarial with the AI agent. These additions increase real-world relevance by adding scale, partial observability, and non-determinism to these task scenarios.

In the updated POGO task, the agent must collect additional resources from additional places (added rooms in Figure 4A) and interact with trading NPCs while avoiding competitive NPCs (seen in Figure 4B). In the updated HUGA task, the agent must navigate varied layouts with different color schemes (Figure 4C). They must also avoid traps, obtain keys to unlock doors and obtain advantages from helpful agents while avoiding adversarial NPCs (seen in Figure 4D).

## PAL Evaluations and Performance

### Running Tournaments in PAL

A tournament manager was developed in python that manages the launching & logging of AI Agents performing in task scenario instances in PAL. This manager can be deployed across computing clusters during evaluations and is parameterized to accept a variety of agents and tournaments to run. The manager runs four separate processes that communicate with each other through sockets on the host machine:

1. PAL, the Minecraft-modded game environment.
2. AI Agent. A bash execution command is necessary to launch the agent.
3. TournamentManager, which monitors instance-ending conditions and sends the appropriate reset command to PAL along with a JSON file describing the next instance to load into PAL.
4. LoggingHandler that logs outputs, records all steps taken, and updates SQL databases with this information. This process includes a tournament timeout – if no progress has been made for more than 5*max game time (i.e., in the event any of the above processes freeze), then the tournament ends.

We provide a copy of the tournament manager that includes processes 1,3, & 4 on the PAL GitHub (https://github.com/PolycraftWorld/PAL). Installation instructions and execution assistance can also be found on the PAL GitHub in the ReadME.txt file. We fully support and highly recommend running this environment in a Debian-based UNIX operating environment (we use Ubuntu 18.04 in our Evaluations); however, it can be successfully deployed in Windows-based environments as well.

AI agents are launched through a user-configurable, UNIX-compliant bash execution command after the PAL environment has been launched and the first task instance has been loaded. Following successful launch, the AI Agent should connect to port 9000 and send an API command to start the tournament.

We have 547 available scenarios across 2 tasks that can be used to train and evaluate an AI agent. Each scenario is a version of the POGO or HUGA task with additional world elements or changes to the arena that provide different challenges to AI agents. These scenarios each have 3 differently seeded tournament variations for a total of 1641 tournaments. Each tournament has 100 games which comes out to 164,100 different game instances. These instances are all available on GitHub (https://github.com/PolycraftWorld/PAL) to train and evaluate on.

## PAL Platform Performance

We are limited in how many actions we can take per second by Minecraft's internal tick rate. Minecraft is hard coded to run at 20 ticks per second and has no setting or configuration available to change the speed. However, this can be unlocked by using reflection to change the tick-rate at runtime. This can be very helpful by increasing training and evaluation times by a factor of 20-40 depending on whether the agent is using visual observations.

We have tested the limits of achievable speed (in ticks per second) in PAL and found that we can reach speeds of 550 ticks per second on modern systems. Our tests were done on a Windows 10 machine with a Ryzen 3900x CPU, 32GB Ram, and a GTX 1070 GPU. This number is mostly affected by single thread processor performance. When an agent is using the screen observation on every tick, this performance degrades significantly. The default screen output format is PNG, which results in an average speed of 73.44 ticks per second. If we change the format of the screen observation to JPEG we get an average speed of 300 ticks per second. We can reach speeds closer to 550 ticks per second if an agent doesn't request the screen after each action (Table S1 in Supplemental Materials).

# Conclusion

Here, we present the Polycraft World AI Lab (PAL), an extensible platform for creating AI challenges, running AI experiments, and evaluating AI agent performance. We have leveraged the strengths of Minecraft but added additional flexibility and accessibility in the PAL system. PAL connects with a broad range of AI agents via our socket-based command interface. Agent evaluation includes comprehensive log files which monitor agents' behavior throughout a task.

We present two examples of custom tasks: a planning task, where the agent is challenged to craft a pogo stick (POGO) and a navigation task, where the agent is challenged to find an item and bring it to a target (HUGA). These tasks were built entirely using the PAL platform. Agents were developed and evaluated on each task. We go on to show that we can add complexity (via added NPCs, task problems, and domain scale) to these tasks to make them more relevant to real-world scenarios. Finally, we discuss the PAL system performance and its ability to rapidly evaluate agents in a tournament structure.

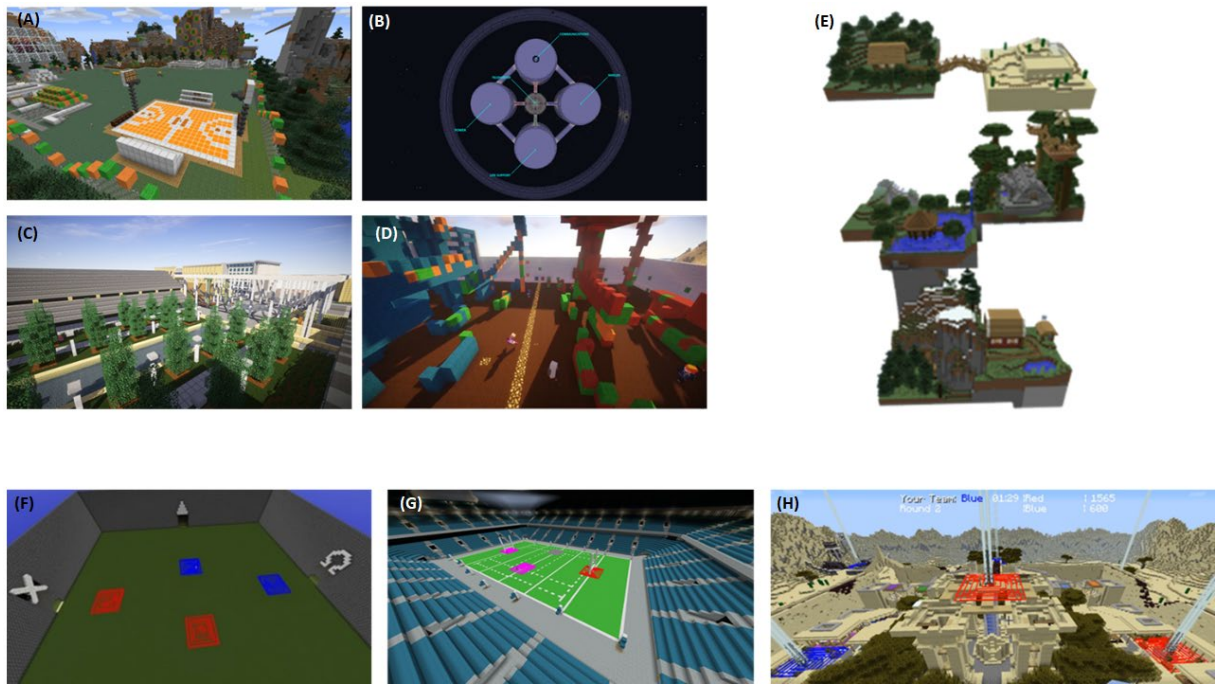

*Figure 5: PAL is capable of a creating a diverse set of tasks. We have created multiple arenas in Polycraft World that have diverse goal conditions. (A) Minigames such as basketball, archery and minigolf (B) An escape room challenge on a space station (C) Scavenger hunts on the UTD campus (D) Virtual paintball (E) a tutorial for teaching gameplay in multiple stages. Our tasks can have various levels of complexity. Capture the base in: (F) a simple square arena, (G) a stadium setting, (H) a desert city.*

This platform is by no means limited to the tasks presented here. We envision this platform will be used to evaluate agents on custom open-world tasks designed by AI researchers and we have demonstrated the capability of the platform to present a wide variety of activities for human and AI agents (Figure 5). We believe the flexibility PAL adds to task design and agent integration are valuable tools for fundamental AI research.

We believe that PAL will enable research on large scale challenges relevant to multi-task learning, transfer learning, and problems encountered in open-world learning including adaptation to out-of-distribution world states and previously unobserved environments (Boult et al., 2021; Kejriwal & Thomas, 2021; Muhammad et al., 2021; Musliner et al., 2021).

Moving forward, this platform will be used for evaluating AI agent performance in novel AI tasks that are increasingly complex, real-world relevant, and involve human-AI teams.

## Conflict of Interest

The authors declare that the research was conducted in the absence of any commercial or financial relationships that could be construed as a potential conflict of interest.

## Author Contributions

S.A.G. was the lead developer for PAL, task creation, manages the project GitHub and wrote the manuscript. R.J.S. wrote the manuscript, was involved in debugging, had several supporting roles and bibliography research. D.N. created the tournament manager. D.V.O. developed the POGO agent, wrote the manuscript and contributed to the project documentation. Y. S. and P. Q. developed the HUGA agent, wrote the manuscript and contributed to the project documentation. J.A. contributed to the project documentation and was involved in debugging. E.K. task creation. E.K., W.V., P.Q. and E.O.V. conceived the project, reviewed the manuscript, and provided guidance and management. All authors reviewed and edited the manuscript.

## Funding

This work was supported in part by the following grant: W911NF2020010 from the Defense Advanced Research Projects Agency. The funding agency is not responsible for the content of this article.

## Acknowledgements

The authors greatly appreciate the constructive feedback of colleagues in the SAIL-ON program for their comments and help testing PAL in the early stages of the platform development. The authors also would like to thank Willie E. Chalmers for the insightful comments and help in the beginning of the project.

# Polycraft World AI Lab (PAL) : An Extensible Platform for Evaluating Artificial Intelligence Agents

## SUPPLEMENTARY INFORMATION

1. Process Communication Standards

Socket communication is key in managing our training and tournament environment. PAL and the AI Agent communicate over port 9000 (default port). The TournamentManager and PAL communicate over port 9005. There is no direct communication between the tournament manager and the AI Agent.

All AI Agent communications **must be initiated by the AI Agent.** PAL will, at no time, send a message unprompted to the AI Agent, instead passing you key information through JSON responses (was your command successfully executed, is the instance over due to a time-out, etc.). In the event that a game timeout does occur, if an agent does not query PAL with a command (like move, or sense_all), the agent will never know that the instance is in fact over and the next instance is waiting to start. Our tournament manager will hang, and eventually, we have a process on our end that will kill the trial and state that the agent has become non-responsive in our logging.

AI Agent Communications are defined by our PAL API, available in the Supplements, in PAL API. In addition to command-specific response variables that vary depending on the command sent, every response JSON contains the following keys:

- *goal*: json dict that contains:
  - *goalType*: "BLOCK_TO_LOCATION" for the HUGA Task, "POGOSTICK" for the POGO Task
  - *goalAchieved*: True if the goal was achieved, False otherwise. Once the goal has been achieved, for the remainder of commands sent in that instance, this will continue to report "True" (boolean)
  - *Distribution*: "Uninformed" if this trial requires "System Detection" of Novelty. "PreNovelty" or "Novelty" if this trial is "Given Detection". (string)
- *command_result*: json dict that contains
  - *command*: command sent by Agent (string)
  - *argument*: any command arguments sent by the Agent (string)
  - *result*: "SUCCESS" if the command was executed properly, "FAIL" if a problem arose in command execution (string)
  - *message*: a non-null string if the result = "FAIL" containing an error message (string)

- o *stepCost*: step cost of executing a particular command (float)
- *step*: Step Number, 0-indexed count of commands sent by the AI Agent (integer)
- *gameOver*: True if the instance is over, False otherwise. (boolean)

Our LoggingHandler monitors the STDOUT and STDERR of the PAL and AI Agent process. This process writes to separate log files all data written to STDOUT and STDERR by the PAL process and by the Agent AI process, broken out as one file per instance.

## 1.1    Communicating Game State to the AI Agent

As indicated above, the tournament manager does not directly communicate details on the state of the trial or the state of the instance to the AI Agent. Instead, the above listed JSON response keys convey critical information that the Agent can process after each command (step) to understand the instance state. As an example, as soon as the goalAchieved=True is found in the response, the Agent knows that it successfully completed the task for that particular instance.

Every JSON response has a "gameOver" key that is either True or False. The value "True" is **sent in only one json response** to the first command received after of the following game-ending criteria is met. Subsequent commands in that same game will revert to having gameOver=False, so it is imperative that the AI Agent recognize and appropriately handle this flag as soon as it appears.

The current instance ending conditions are:

1. Successful Completion of Task (we also include another key, "goalAchieved" in each response that will reflect this case)
2. Game Time limit exceeded
3. Total Step Cost exceeds a step-cost maximum of 1,000,000 (an empirically determined upper bound to prevent abuse of API commands or other counter-productive behavior towards task completion)
4. Agent sends the GIVE_UP command

As of now, our architecture requires that the AI agent, following the successful reception of a response JSON containing gameOver=True send one last command back to PAL to acknowledge receipt and initiate the reset to start the next instance (doesn't matter what the command is, as its step cost is not counted for/against your score, and its response JSON is irrelevant for your agent as well. Most agents just pass a SENSE_ALL command and ignore the response). Ideally, only one additional command will be processed by PAL following any of the instance end conditions evaluating to true.

Tournament completion is not directly communicated towards the AI Agent. Instead, the process is terminated by the parent process.

## 1.2 Running Tests of your Agent in PAL

Detailed Instructions on setup and installation of the Tournament Manager environment is available on GitHub (https://github.com/PolycraftWorld/PAL). To run an AI Agent, the AGENT_COMMAND config variable (or command line parameter) needs to be adjusted to point to a script that will load the AI Agent. As mentioned above, the AI agent is expected to, as part of that startup script, try to connect to port 9000 and send an API command to begin the trial. Though an AI agent could just be a folder of scripts as our baseline agent is, we found that many of our partners prefer to run their AI agents in Docker containers – that works well in our current evaluation environment, too.

## 1.3 Handling Updates to PAL in your Test Environment

In the event that updates or bug-fixes are made to PAL (I.e., new API commands, new novelties, etc.), we will communicate via email/slack/teams with a detailed changelog. Please direct your code to the appropriate branch on PAL GitHub (https://github.com/PolycraftWorld/PAL), pull the updates, and then continue to run tests using your AI agent.

2. Example of Command and Response

An agent can use the API by connecting to the default port 9000 using the following Python script:

```python
import socket
import json

# connect to socket
HOST = '127.0.0.1'
PORT = 9000
sock = socket.socket(socket.AF_INET, socket.SOCK_STREAM)
sock.connect((HOST, PORT))
```

The following script constructs a function (MC) that enables the communication with Polycraft:

```python
def MC(command):
    "function that enable the communication with minecraft"
    print( command )
    sock.send(str.encode(command + '\n'))
    BUFF_SIZE = 4096  # 4 KiB
    data = b''
    while True:
        part = sock.recv(BUFF_SIZE)
        data += part
        time.sleep(.001)
        if len(part) < BUFF_SIZE or part[-1] == 10:
            # either 0 or end of data
            break
    print(data)
    data_dict = json.loads(data)
    # print(data_dict)
    return data_dict
```

Now we can start the API using :

```python
MC('START')
```

And reset the task:

```python
MC("RESET                                                          domain
../pogo_100_PN/POGO_L00_T01_S01_X0100_U9999_V0_G00000_I0020_N0.json")
```

Now we can send commands. For example:

```python
MC('SENSE_RECIPES'))
```

```
MC('SENSE_ALL')
MC('SELECT_ITEM minecraft:iron_pickaxe')
```

After the command is processed, a text JSON response will be sent back which includes the result of the command and task goal information. For example, the return of the last command will be:

```json
{
 "goal": {
 "goalType": "ITEM",
 "goalAchieved": false,
 "Distribution": "Uninformed"
 },
 "command_result": {
                         "command": "select_item",
                         "argument": "minecraft:iron_pickaxe",
                         "result": "SUCCESS",
                         "message": "selected item",
                         "stepCost": 120
                         },
 "step": 0,
 "gameOver

 }
```

3. PAL API

The PAL API consists of different API commands broken down into SYSTEM commands, DEV commands, and GAME commands. For AI Agent testing using the TournamentManager, only GAME commands will need to be used. The other commands are provided for completeness only.

3.1 Available configuration

There are three places available for configuration settings for a tournament run. There are the command-line arguments, the config file (PAL/PolycraftAIGym/config.py), and lastly there are available environment variables.

### 3.1.1 *LaunchTournament.py Command-line arguments*

**LaunchTournament.py Usage:**

LaunchTournament.py will start a new tournament on the system. It will first start a Polycraft client, then start a tournament manager process to initialize the first game of the tournament. Once the game has initialized, it will start the agent.

**Format**:

LaunchTournament.py **-c** <game_count> **-t** <tournament_name> **-g** <game_folder> **-a** <agent_name> **-d** <agent_directory> **-x** <agent_command> **-i** <maximum time (sec)> **-m** <max tournament time (minutes)>

**-c** How many games to run.

 Default: 100

**-t** Name of tournament.

 Default: "POGO_L00_T01_S01_X0100_A_U9999"

**-g** Where games are located.

 Default: "../pogo_100_PN"

**-a** Name of your agent.

 Default: "MY_AGENT_ID"

**-d** Where agent is located

 Default: ""

**-x** Command to run agent. Text as would be entered into command-line. To escape quotes or other special characters, use "\"

 Default: "python pogo_agent.py"

**-i** Seconds per game

 Default:                 300

**-m** Max time for tournament to run in minutes

 Default: 2880

**Example:** python LaunchTournament.py -c 100 -t "POGO_L00_T01_S01_X0100_A_U9999_V0200FPS_011022" -g "../pogo_100_PN" -a "BASELINE_POGOPLAN " -d "agents/pogo_stick_planner_agent/" -x "python python_miner_PLANNER.py"

**PAL Observation Variables**

**SENSE_SCREEN_FORMAT** – Some screen formats will compress faster than others at the expense of quality. PNG seems to give the best quality, while JPEG typically give the best performance. The default image type is PNG but this can be set to the following: PNG, BMP, JPEG, JPG, WBMP, GIF.

**AIGYM_REPORTING**  - Set this to "True" to activate AIGym style reporting. The agent will receive a "SENSE_ALL" output after every command.

**REPORT_SCREEN** – If "AIGYM_REPORTING" is set to true, you can also set this to "True" to also receive a "SENSE_SCREEN" output after every command.

**PAL Ports**

**PAL_AGENT_PORT** – Set the agent socket communication port. Default is 9000. **PAL_TM_PORT** – Set the tournament manager socket communication port. Default is 9005.

**PAL Speed**

**PAL_FPS** – Set the frames per second for PAL to run. Default is 20.  The maximum value here is 1000, but most systems cannot achieve that unless they have powerful single thread cores and a powerful GPU.

Internal testing revealed some results you may be able to expect on your own machine.

**Sense screen testing**

Test System specs:
OS – Windows 10 Pro
CPU – Ryzen 9 3900x 12 core @4.00 GHz
RAM                                                            –                                                            32GB
GPU – GTX 1070

**Table S1 – Max ticks per second.**

| Target ticks per second | Tested ticks per second (with different screen output formats) | | | | | |
|---|---|---|---|---|---|---|
| | PNG | BMP | JPG | WBMP | GIF | NONE |
| 20 (Default) | Min:18.79 Max:20.07 Mean:19.98 Median:20.0 | Min:18.68 Max:20.08 Mean:19.99 Median:20.0 | Min:19.0 Max:20.1 Mean:19.99 Median:20.0 | Min:19.04 Max:20.18 Mean:19.99 Median:20.0 | Min:19.0 Max:20.06 Mean:19.99 Median:20.0 | Min:18.49 Max:20.09 Mean:19.98 Median:20.0 |
| 200 | Min:51.4 Max:145.92 Mean:78.54 Median:74.91 | Min:95.71 Max:174.4 Mean:127.56 Median:127.36 | Min:105.71 Max:187.93 Mean:153.39 Median:156.33 | Min:98.24 Max:205.07 Mean:176.07 Median:179.92 | Min:99.23 Max:188.43 Mean:123.46 Median:114.71 | Min:169.78 Max:209.27 Mean:202.42 Median:201.09 |
| 500 | Min:42.48 Max:93.44 Mean:56.62 | Min:72.5 Max:250.81 Mean:148.03 | Min:120.09 Max:349.22 Mean:263.06 | Min:101.79 Max:458.81 Mean:305.63 | Min:90.22 Max:199.1 Mean:142.96 | Min:340.2 Max:511.38 Mean:470.79 |
| | Median:53.59 | Median:127.73 | Median:262.68 | Median:406.73 | Median:192.48 | Median:419.29 |
| 800 | Min:50.3<br>Max:151.18<br>Mean:73.44<br>Median:77.24 | Min:130.46<br>Max:277.02<br>Mean:227.44<br>Median:130.46 | Min:121.36<br>Max:409.22<br>Mean:297.09<br>Median:121.36 | Min:97.65<br>Max:384.09<br>Mean:213.5<br>Median:176.36 | Min:108.61<br>Max:196.33<br>Mean:155.94<br>Median:108.61 | Min:392.07<br>Max:677.97<br>Mean:549.36<br>Median:539.08 |
| 1000 | Diminishing performance past 800 | | | | | |

**PAL Screen Position**

**PAL_X** – x position on screen for window to render. Default is chosen by Minecraft

**PAL_Y** – y position on screen for window to render. Default is chosen by Minecraft

**PAL_WIDTH** – Screen render width size. Default is 256

**PAL_HEIGHT** – Render screen height size. Default is 256

3.2 GAME commands

### 3.2.1 Tournament Commands

These commands are sent by the AI Agent to communicate directly to the Tournament Manager and should be used when the Agent is being officially evaluated, where appropriate.

- **GIVE_UP**
  - Agent gives up task, letting tournament manager know that the next instance can be queued for a domain reset.
  - Please note that following a GIVE_UP command, the agent will receive a gameOver=True as part of the response JSON. The Agent is still required to send one additional command to PAL to acknowledge receipt of the gameOver=True before the RESET domain command is triggered by the Tournament Manager and the next instance in the trial is loaded
- **REPORT_NOVELTY** [–l lvl] [–c confidence] [–m user-message]
  - indicates that you have detected novelty with optional parameters | [-l novelty level]
  - [-c confidence interval 0f:100f] [-m user-defined message]

### 3.2.2 Movement commands

These commands enable the AI Agent to move around the instance.

- **MOVE** w|a|d|x
  - moves 1 meter forward (w), left (a), right (d) or back (x) (i.e. "MOVE w" moves forward 1 block)
- **MOVE** q|e|z|c
  - moves sqrt (2) distance diagonally with q,e,z,c (i.e., "MOVE q" moves diagonally forward and leftward, relative to the player's facing direction). Diagonal movements correspond to relative location of letter against the WSAD keys.
- **TURN** -15
  - alters player's horizontal facing direction (yaw) in 15-degree increments (no interpolation)
  - The parameter passed must be a multiple of 15, positive or negative.
- **SMOOTH_TILT** FORWARD

- o Sets the player's pitch to the horizon (0 degrees)
- o While this is a "SMOOTH" command, **it operates as an abrupt state transition.** There are plans to create a "TILT" command once the visual interpolation version of the command is deployed.
- **SMOOTH_TILT** DOWN
  - o Sets the player's pitch to -45 (looking directly at the ground in front of the player – ideal for viewing the location upon which a block should be placed on the ground)
  - o While this is a "SMOOTH" command, **it operates as an abrupt state transition.** There are plans to create a "TILT" command once the visual interpolation version of the command is deployed.
- **TP_TO** x,y,z [distance]
  - o Teleports the player to the block at x,y,z, resetting the player's pitch and yaw to have the player face North and set their pitch to 0 degrees (horizon), positioning the player a default distance of 1 away from the block if no distance is passed. In other words, if distance is Null/not given, the player's true position would be [x,y,z-1] and executing a MOVE W command will place the player onto the exact coordinate.
  - o The optional distance parameter must be a positive integer greater than or equal to 1. This distance will adjust the resulting position of the player to be [x,y,z-distance], facing the coordinate desired at an offset [distance] blocks away. Running the MOVE W command [distance] number of times will place the player on the target coordinate [x,y,z].
  - o Using distance = 2 is necessary when playing tree-taps on trees, as otherwise, the player will be occupying the space that the tree-tap will need to occupy after being placed.
  - o The distance offset must yield an allowable move_to location (i.e., it can't be a block in the area) or command fails.
- **TP_TO** entityID
  - o teleports to the location of an entity with entity_ID = entityID (i.e., TP_TO 7101 teleports to an entity with the ID "7101")

### 3.2.3   Sensing commands
- **CHECK_COST**
  - o returns the total stepCost incurred since the last RESET command
- **SENSE_INVENTORY**
  - o returns contents of player inventory in .json format
- **SENSE_LOCATIONS**
  - o returns sensible world environment (blocks, entities and locations) as .json
- **SENSE_RECIPES**
  - o Returns the list of recipes available in the experiment
- **SENSE_SCREEN**
  - o Returns pixels sent to the display output window, in the form of a string listing an array of integers
- **SENSE_ALL**
  - o returns inventory, recipe and information on all observable blocks around the player's position.
- **SENSE_ALL NONAV**
  - o returns inventory, recipe and location information in .json
  - o NONAV parameters omits information which is not needed for agents that do not navigate the world

### 3.2.4   Interacting commands

- **SELECT_ITEM** polycraft:wooden_pogo_stick
  - sets a specific item from your inventory in your hand as the active item (e.g. tool or block)
- **USE_HAND**
  - to open doors with bare hand (ignores item in hand to interact)
- **BREAK_BLOCK**
  - breaks block directly in front of player with selected item
  - selected item and block type yield stepCost of action
- **CRAFT** 1 minecraft:log 0 0 0
  - note that CRAFT must be followed by a "1"
  - crafts 4 Planks (uses the player's personal 2x2 crafting grid in their inventory)
- **CRAFT** 1 minecraft:planks 0 minecraft:planks 0
  - crafts 4 Sticks (uses the player's personal 2x2 crafting grid in their inventory)
- **CRAFT** 1 minecraft:planks 0 0 minecraft:planks 0 0 0 0 0
  - crafts 4 Sticks using a crafting table (3x3 grid)
- **CRAFT** 1 minecraft:stick minecraft:stick minecraft:stick minecraft:planks minecraft:stick minecraft:planks 0 polycraft:sack_polyisoprene_pellets 0
  - crafts a Wooden Pogo Stick on a crafting table
- **EXTRACT_RUBBER**
  - moves polycraft:sack_polyisoprene_pellets from polycraft:tree_tap to player inventory
- **PLACE_TREE_TAP**
  - calls PLACE_BLOCK polycraft:tree_tap (and processes extra rules)
- **PLACE_CRAFTING_TABLE**
  - calls PLACE_BLOCK minecraft:crafting_table (and processes extra rules)
- **PLACE_MACGUFFIN**
  - calls PLACE_BLOCK polycraft:macguffin (and processes extra rules)

### 3.2.5   Game commands

These commands should be used during developmental testing only. When testing the AI Agent using the TournamentManager, these commands are managed by the main Python thread and **should not be called by the AI Agent.**

- **START**
  - no args ever used | called once to start trials
- **RESET** domain ../available_tests/pogo_nonov.json
  - the base pogo experiment
- **RESET** domain ../available_tests/hg_nonov.json
  - the base hunter-gatherer experiment

### 3.2.6   Dev commands

Dev commands must be enabled by setting a client virtual machine argument: "-Ddev=True" Details on setting this outside of a development environment are still being worked out, as solutions are fickle and system dependent. Please contact us if you need these commands.

- **CHAT** "Hello world."
- **CHAT** /give @p minecraft:stick

- o   not used in DRY-RUN Tournaments, but active for debugging/training/development
- **TELEPORT** 20 4 21 90 0
  - o   not to be used in DRY-RUN Tournaments, but allows setting player location and view direction.
  - o   Parameters: [x] [y] [z] [yaw] [pitch]